\documentclass[sigconf]{acmart}
\usepackage{subcaption}
\usepackage{siunitx}
\DeclareMathOperator\atanh{atanh}

\AtBeginDocument{%
  \providecommand\BibTeX{{%
    \normalfont B\kern-0.5em{\scshape i\kern-0.25em b}\kern-0.8em\TeX}}}

\copyrightyear{2022}
\acmYear{2022}
\setcopyright{acmlicensed}\acmConference[GECCO '22]{Genetic and
	Evolutionary Computation Conference}{July 9--13, 2022}{Boston, MA, USA}
\acmBooktitle{Genetic and Evolutionary Computation Conference (GECCO '22),
	July 9--13, 2022, Boston, MA, USA}
\acmPrice{15.00}
\acmDOI{10.1145/3512290.3528721}
\acmISBN{978-1-4503-9237-2/22/07}

\begin{document}

\title{Hyperparameter Tuning in Echo State Networks}

\author{Filip Matzner}
\email{filip.matzner@mff.cuni.cz}
\orcid{0000-0001-5996-8794}
\affiliation{%
  \institution{Charles University\\
  Faculty of Mathematics and Physics}
  \streetaddress{Malostranské náměstí 25}
  \city{Prague}
  \country{Czech Republic}
  \postcode{118 00}
}


\begin{abstract}
Echo State Networks represent a type of recurrent neural network with a large randomly generated reservoir and a small number of readout connections trained via linear regression.
The most common topology of the reservoir is a fully connected network of up to thousands of neurons.
Over the years, researchers have introduced a variety of alternative reservoir topologies, such as a circular network or a linear path of connections.
When comparing the performance of different topologies or other architectural changes, it is necessary to tune the hyperparameters for each of the topologies separately since their properties may significantly differ.
The hyperparameter tuning is usually carried out manually by selecting the best performing set of parameters from a sparse grid of predefined combinations.
Unfortunately, this approach may lead to underperforming configurations, especially for sensitive topologies.
We propose an alternative approach of hyperparameter tuning based on the Covariance Matrix Adaptation Evolution Strategy (CMA-ES).
Using this approach, we have improved multiple topology comparison results by orders of magnitude suggesting that topology alone does not play as important role as properly tuned hyperparameters.
\end{abstract}




\maketitle

\section{Introduction} \label{sec:introduction}

Recurrent neural networks provide an efficient way of data processing in organic tissue.
However, training artificial recurrent networks still poses a challenge~\cite{pascanu2012difficulty} even though the artificial models are vastly simplified compared with their biological counterparts.
In order to overcome the difficulty of training and avoid vanishing and exploding gradient problems, \citeauthor{jaeger2001echo}\cite{jaeger2001echo} proposed a model called \textit{Echo State Network} (ESN).
The model uses a large recurrent reservoir which stays fixed throughout the whole lifetime of the network and, therefore, completely avoids the gradient propagation pitfalls. ESNs are especially useful for time series modelling~\cite{schrauwen2007overview}.
In real-world applications, they have demonstrated promising results e.g., in financial markets~\cite{kim2020prediction}\cite{crisostomi2015electric}, weather forecasting~\cite{kim2020prediction}\cite{mcdermott2017weather}\cite{yen2019rainfall}, microsleep detection~\cite{ayyagari2015leaky} and many more.



\begin{figure}
  \centering
  \begin{subfigure}[t]{0.32\linewidth}
    \includegraphics[width=\textwidth]{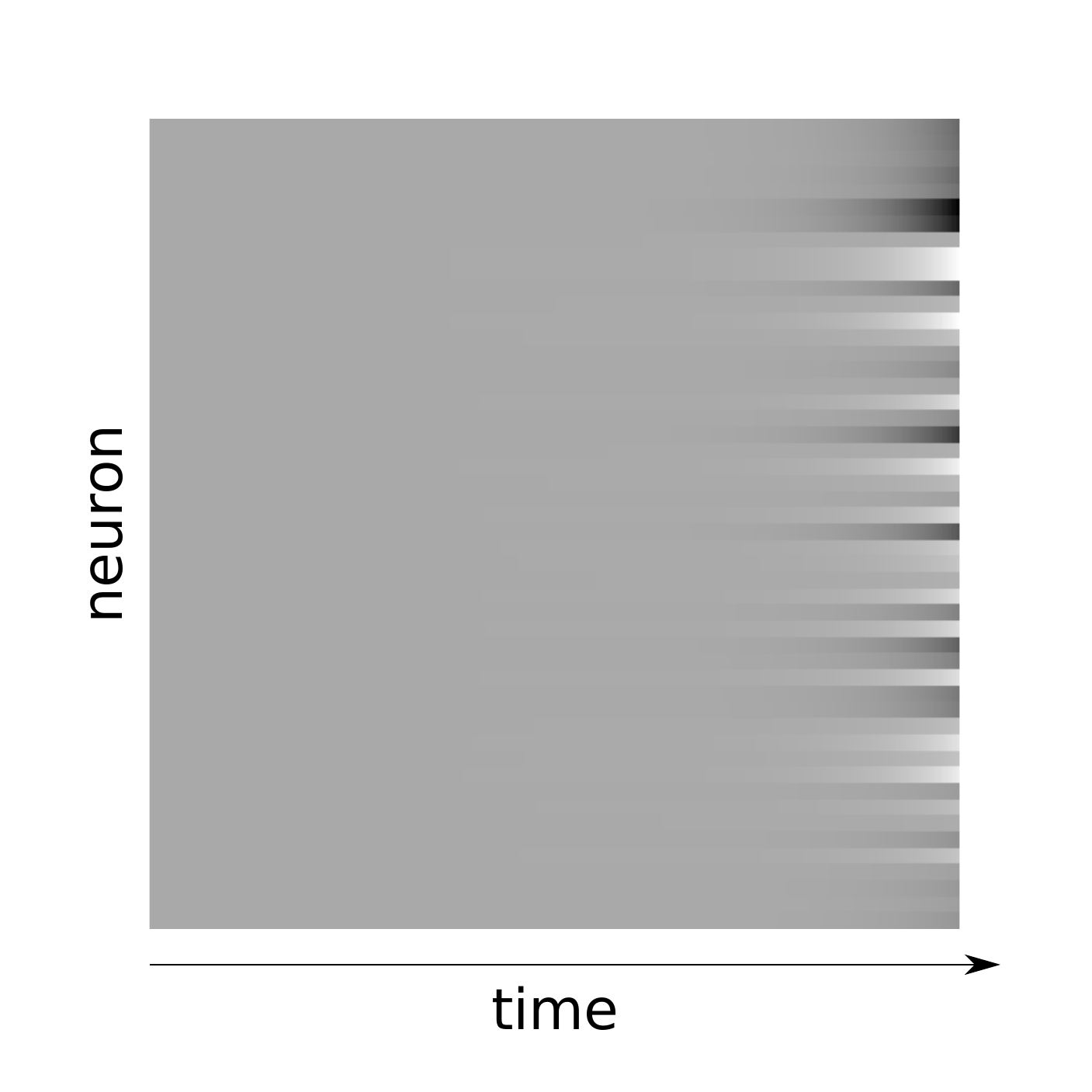}
    \captionsetup{width=0.8\textwidth}
    \caption{Ordered net activity converges to a fixed state.}
    \label{fig:ordered_net}
  \end{subfigure}
  \begin{subfigure}[t]{0.32\linewidth}
    \includegraphics[width=\textwidth]{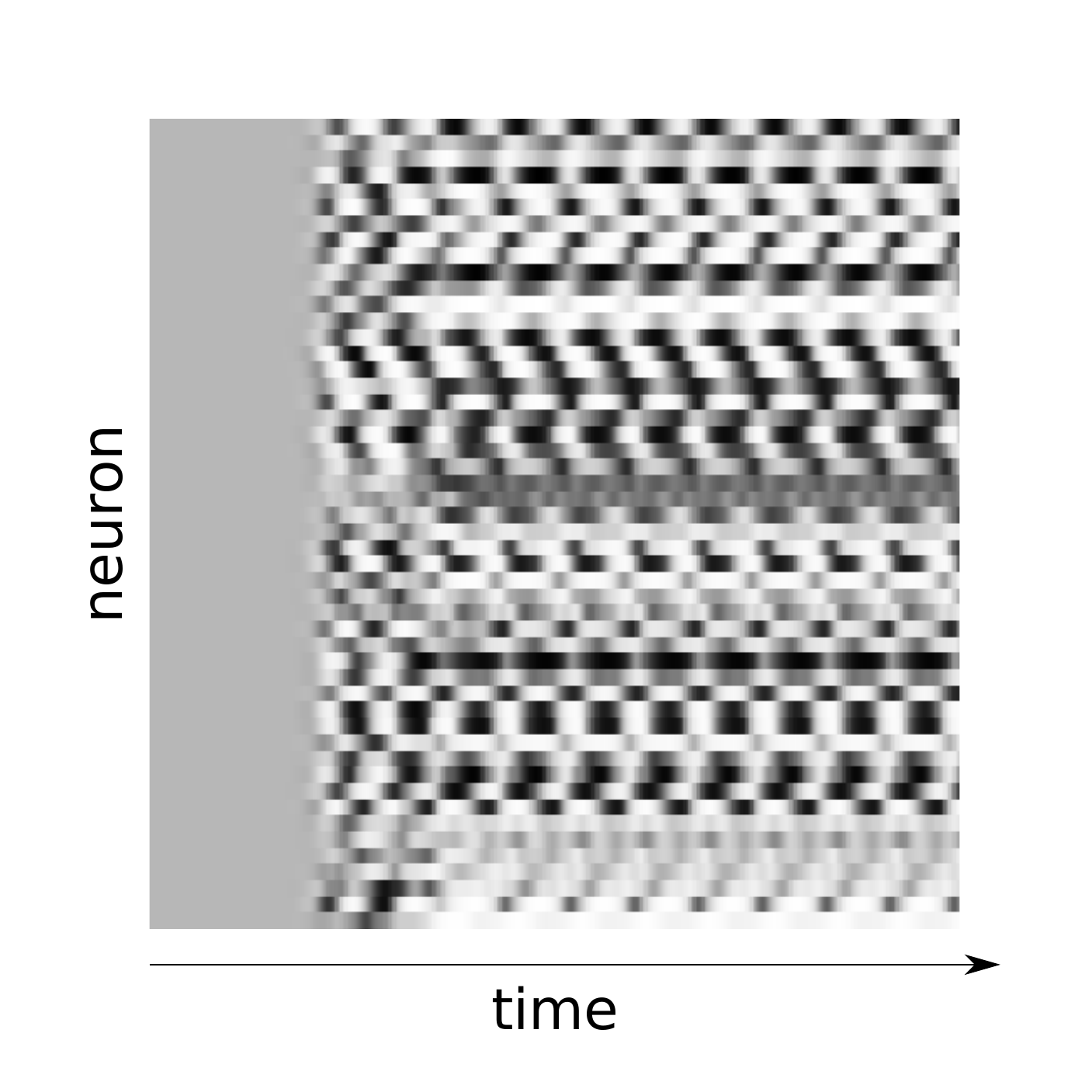}
    \captionsetup{width=0.8\textwidth}
    \caption{Edge of chaos net produces repeating patterns.}
    \label{fig:edge_of_chaos_net}
  \end{subfigure}
  \begin{subfigure}[t]{0.32\linewidth}
    \includegraphics[width=\textwidth]{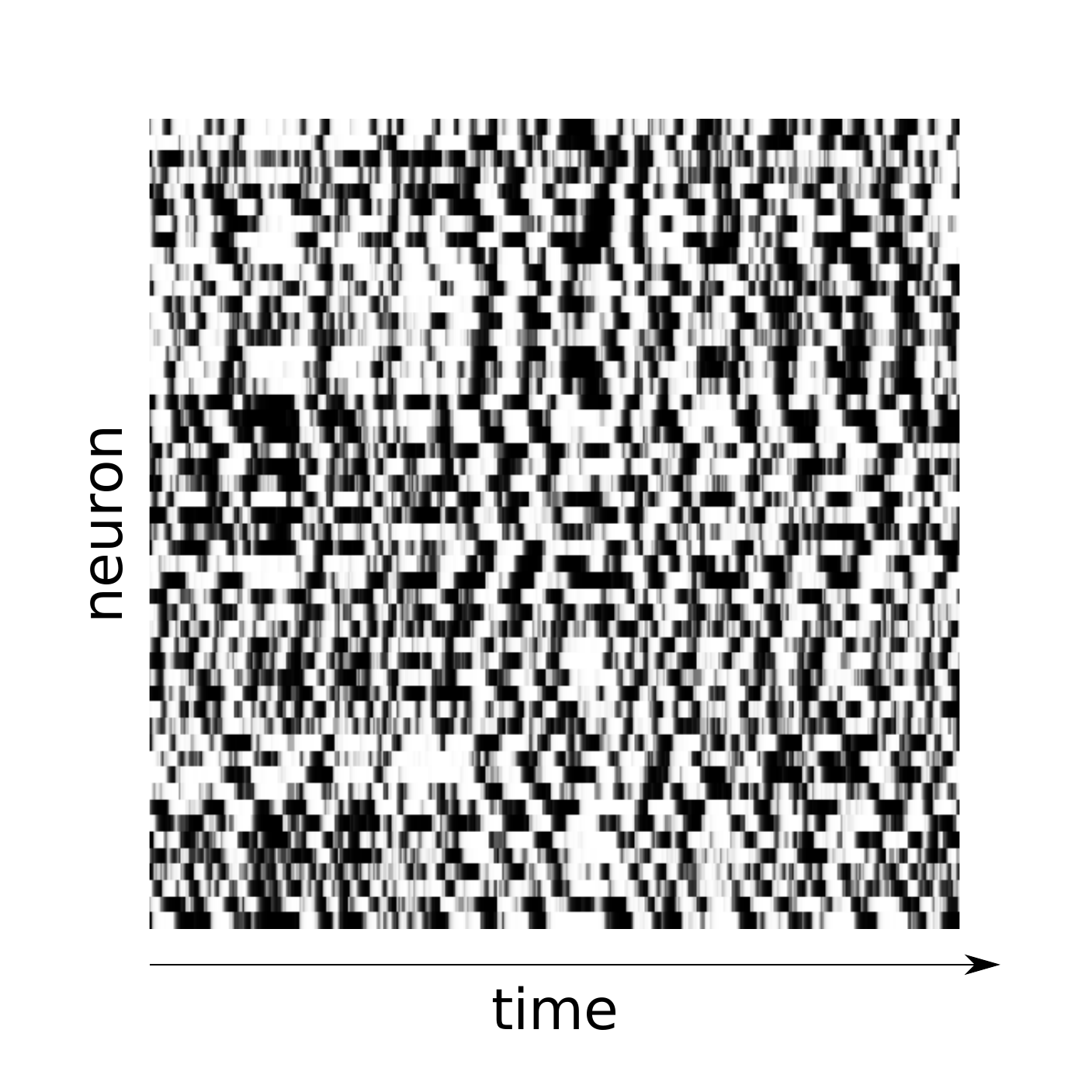}
    \captionsetup{width=0.8\textwidth}
    \caption{Chaotic net output resembles random white noise.}
    \label{fig:chaos_net}
  \end{subfigure}
  \caption{The states of three echo state networks tuned for ordered, edge of chaos and chaotic dynamics (left to right). Each column represents the activation of all its 50 reservoir neurons. The networks were driven by one positive input and the rest of the inputs were set to zero.}
  \label{fig:three_dynamics}
  \Description{Three different images each one containing a visualization of an echo state network in one of the dynamic regimes - ordered, edge of chaos, and chaotic.}
\end{figure}

ESNs are specified by a set of hyperparameters that define the behavior and properties of the network.
Even though they are claimed to be robust against various hyperparameter changes~\cite{jaeger2001echo}, we will demonstrate that improper tuning of those hyperparameters can severely impact the network’s performance. The most usual method of ESN hyperparameter tuning is a \textit{grid search} \cite{jaeger2002tutorial}\cite{lukosevicius2009approaches}\cite{maat2018optimization}\cite{lukosevicius2012practical},
where the best hyperparameters are chosen from a predefined set of combinations.
Unfortunately, this method may lead to underperforming networks and, therefore, questionable results when comparing different topologies or other architectural changes.

According to some authors (e.g.,~\cite{matzner2017neuroevolution}~\cite{bertschinger2004real}), ESNs tend to maximize their performance when their recurrent dynamics is close to the so-called \textit{edge of chaos}.
Unfortunately, networks in this regime are very sensitive to hyperparameter changes and even a small perturbation can turn a flawless network to a chaotic white noise generator (Figure~\ref{fig:three_dynamics}).
Besides the overall complicated hyperparameter space of a recurrent system, this nonlinear sensitivity adds in one more challenge that makes hyperparameter tuning difficult.

Contrary to e.g., \textit{backpropagated} neural networks, the ESN training procedure is very fast.
This fundamental property allows for various hyperparameter optimization techniques requiring thousands of \textit{fitness} evaluations, which would be very impractical for traditional neural network models.
With that in mind, we propose a tuning method based on an \textit{Covariance Matrix Adaptation Evolution Strategy (CMA-ES)}~\cite{hansen2001cma} for continuous optimization, which has the potential to demonstrate the real possibilities of echo state networks.

One of the downsides of the proposed tuning algorithm are its computational demands.
We need to run the full network evaluation thousands of times to make sure we converge to an optimum.
For such a time consuming procedure to finish in a reasonable time, we have created a highly optimized C++ implementation of ESNs for graphical processors (GPU).
Nevertheless, each of our many experiments still consumes tens of hours of computational time on a consumer-grade CPU and GPU.

The goal of the experiments presented in this paper is not to present a time-efficient method of ESN optimization, but to prove the existence of unexpected and performant parameter configurations difficult to find using a grid search or manual tuning.
The experiments will compare the results of various ESN improvements found in literature with a properly tuned basic model. Narrowing the search space or reducing the required computational resources is left for future work.

\section{Related Work}

One of the downsides of CMA-ES is its requirement of large number of function evaluations to be able to provide satisfying results.
That is why its utilization in models with long evaluation times, such as deep neural networks, may be challenging.
There are alternative approaches promising lower number of evaluations, usually with a performance trade-off.
Some of the examples are Bayesian optimization~\cite{maat2018optimization}\cite{yperman2016bayesian}\cite{cerina2020echobay}, random search~\cite{bergstra2012random}, gradient based methods \cite{thiede2012gradient}, evolutionary algorithms~\cite{wang2015swarm}\cite{matzner2017neuroevolution}\cite{dale2018neuroevolution}, and even analytical approaches~\cite{livi2017edge}. There are also attempts to speed up the CMA-ES itself by warm starting it using previous or transferred knowledge~\cite{nomura2020warm}.

Despide the challenges, there has been multiple attempts to use CMA-ES for hyperparameter tuning of ESNs and neural networks in general.
For instance, \citeauthor{loshchilov2016hyperparameter}~\cite{loshchilov2016hyperparameter} used CMA-ES for hyperparameter optimization of deep neural networks on handwritten digit recognition task.
With sufficient number of evaluations and enough computational time, CMA-ES eventually outperformed state-of-the-art Bayesian methods.
In a recent study, \citeauthor{liu2020nonlinear}~\cite{liu2020nonlinear} employed CMA-ES specifically in echo state network hyperparameter optimization and reached similar results.
However, even though their model outperformed other optimization approaches in the three tasks evaluated in the study, their approach had a few drawbacks preventing it from reaching state-of-the-art performance.
First, only three hyperparameters were optimized, crucially restricting the network's possibilities.
Second, the training data had only 1000 time steps, which seems inadequate for a reservoir of up to 1000 neurons (discussed in Section~\ref{sec:discussion}).


In our paper, the list is extended to eight hyperparameters, which on one hand provides more degrees of freedom, but on the other hand, the extended search space and its sensitivity pose a challenge for the optimizer.
Furthermore, some hyperparameters require careful steps as their absolute value closes to zero while some prefer linear steps throughout their whole domain.
To address the issue, we have transformed the search space of some parameters according to their needs.

\section{Methods}

\subsection{Echo State Networks}

\begin{figure}
    \centering
    \includegraphics[width=\linewidth]{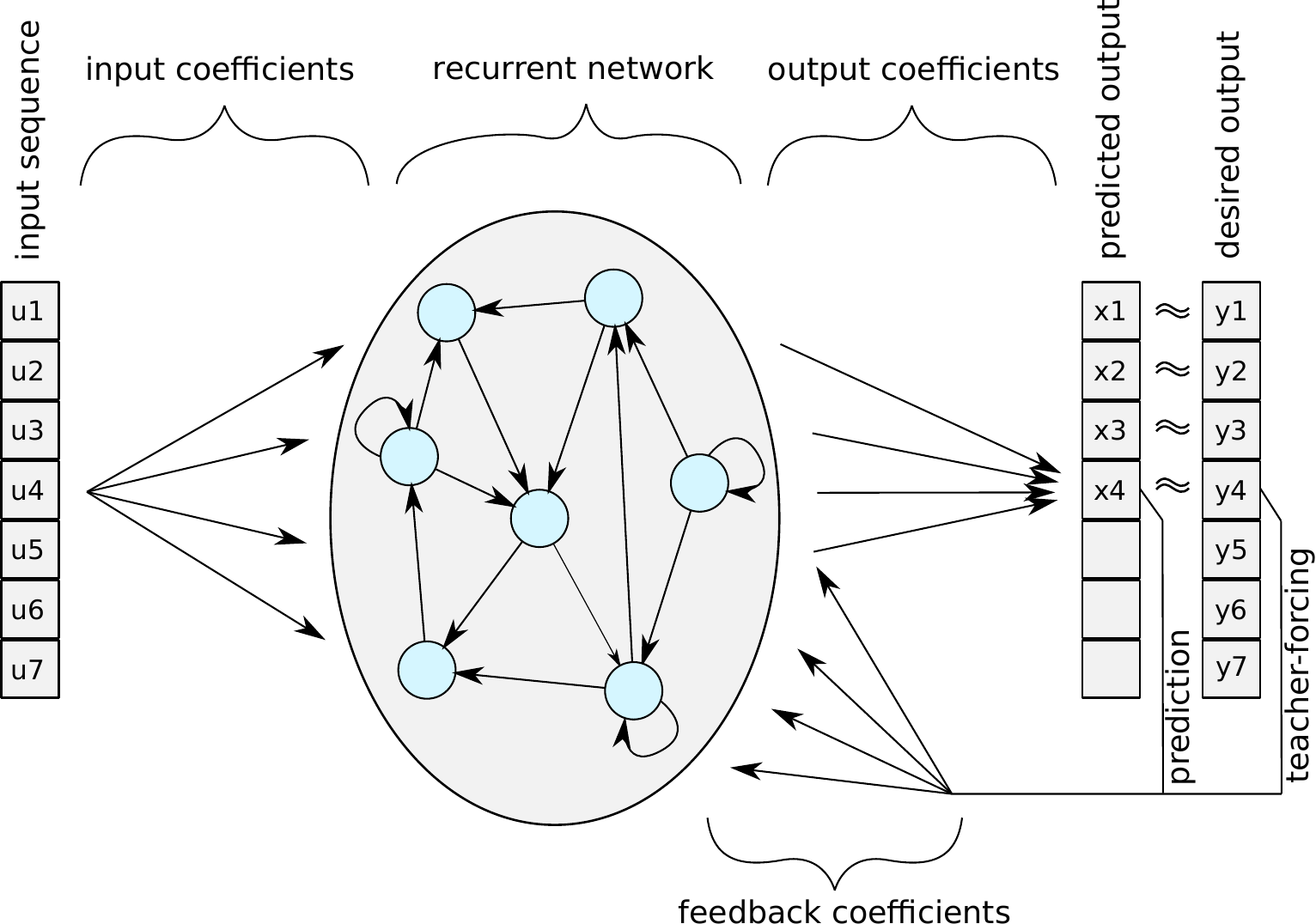}
    \caption{Echo state network overview.}
    \label{fig:esn_intro}
    \Description{The individual layers of an echo state network. Input weights, reservoir, readout layer.}
\end{figure}

An echo state network (Figure~\ref{fig:esn_intro}) with $n$ neurons as defined by \citeauthor{jaeger2001echo}~\cite{jaeger2001echo} consists of a \textit{reservoir} with connectivity matrix $W \in \mathbb{R}^{n \times n}$, a vector of \textit{input coefficients} denoted by $w_{\textrm{in}} \in \mathbb{R}^{n}$, a vector of \textit{readout coefficients} denoted by $w_{out} \in \mathbb{R}^{n}$ and a vector of \textit{feedback coefficients} denoted by $w_{fb} \in \mathbb{R}^{n}$.
The activations of the neurons in the recurrent network in time $t$ are denoted as $a(t) \in \mathbb{R}^{n}$, the input value as $u(t) \in \mathbb{R}$, the output value as $x(t) \in \mathbb{R}$ and the desired output as $y(t) \in \mathbb{R}$. The recurrence and the output are calculated as follows:
 \begin{align*}
    \textstyle
    z(t) &= W a(t-1) + w_{in} u(t) + w_{fb} x(t-1) + \mu_{b} \,, \\
    a(t) &= (1 - \gamma) a(t-1) + \textrm{tanh}(z(t)) \odot \textrm{N}_{n}(1, \epsilon^2) \,, \\
    x(t) &= a(t) \cdot w_{out} \,,
    \label{eq:esn}
 \end{align*}
where $z(t) \in \mathbb{R}^{n}$ are the raw potentials before being processed by $\textrm{tanh}$ activation function, $\gamma \in \mathbb{R}$ is the leakage parameter, $\mu_{b} \in \mathbb{R}$ denotes the constant bias, $\odot$ represents elementwise multiplication, $\text{N}_{n}(1, \epsilon^2) \in \mathbb{R}^{n}$ is a vector with each element drawn independently from \textit{normal distribution} $\text{N}(1, \epsilon^2)$, and $\epsilon \in \mathbb{R}$ denotes the strength of the internal noise.
 
The reservoir matrix and the input and feedback coefficients are generated randomly and stay fixed. The output coefficients $w_{out}$ are trained using linear regression so that the predicted sequence $x(t)$ and the desired output sequence $y(t)$ minimize their squared distance. Specifically, during the training, the network is driven by the training input sequence $u$ with the desired output sequence $y$, both of length $k$. Let us define matrix $A \in \mathbb{R}^{k \times n}$ whose i-th row is the vector of activations $a(i)$. Afterwards, the least squares method is used to find $w_{out}$ as follows:
\begin{equation*}
    \textstyle
	w_{out} = \arg\min_{w} \left| \left| Aw - y \right| \right| \,,
	\label{eq:esn_lincoef_lstsq}
\end{equation*}
where $|| \cdot ||$ denotes the \textit{Euclidean norm}.

During the training phase (i.e., before the initialization of $w_{out}$), the feedback connections are driven by the desired output sequence $y(t)$ and during the testing phase, the feedback connections use the network prediction $x(t)$.
This process is known as teacher-forcing~\cite{jaeger2001echo}\cite{lukosevicius2012practical} and it is depicted in Figure~\ref{fig:esn_intro}.

\subsection{Topologies}

\begin{figure}
    \centering
    \includegraphics[width=\linewidth]{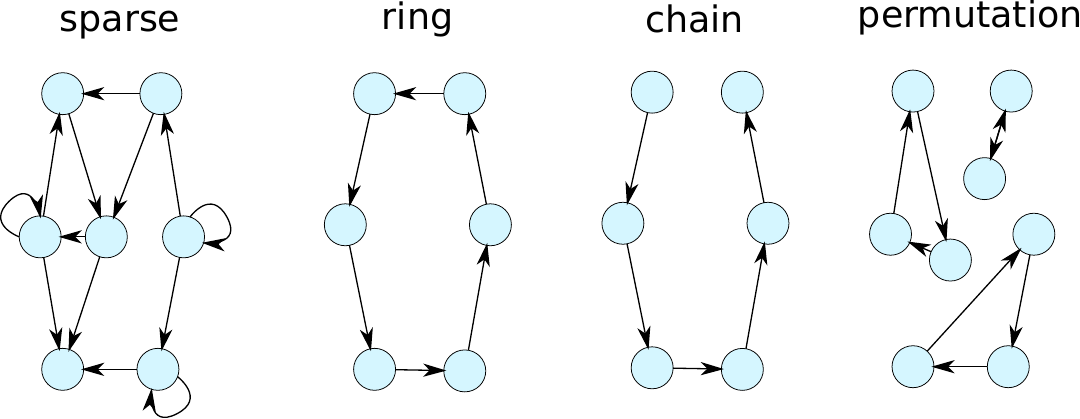}
    \caption{Common ESN topologies.}
    \label{fig:topologies}
    \Description{The tested topologies - sparse, permutation, ring, and chain.}
\end{figure}

In our experiments, we compare four variants of the reservoir as depicted in Figure~\ref{fig:topologies}.
The most common variant is called \textit{sparse} topology.
Its reservoir is a fully connected network where every edge has a probability $\alpha \in [0, 1]$ to be missing.
This corresponds to a weight matrix $W$ initialized from $\text{N}(\mu_{res}, \sigma_{res}^2)$ whose elements are set to zero with the probability $\alpha$.

Another usual topology is called chain. It is created by numbering the neurons from $1$ to $n$ and connecting neuron $i$ to neuron $i+1$ while leaving the last neuron $n$ disconnected. The corresponding weight matrix is shown in Equation~\ref{eq:chain}.

\begin{equation}
\textstyle
\begin{matrix}
    0 & \dots  & \dots  & 0 \\
    w_{1} & \ddots   &    & \vdots \\
    \vdots &  \ddots & \ddots & \vdots \\
    0 & \dots  & w_{n}  & 0
    \label{eq:chain}
\end{matrix}
\end{equation}

If the last neuron $n$ connects back to neuron $1$, it is called the \textit{ring} topology.
The reservoir matrix of the ring is similar to the chain reservoir, but it has a nonzero value $w_{n+1}$ in its upper right element.
Both the chain and ring topologies have been shown to have similar properties as the sparse topology with regard to memory capacity and some authors have suggested that the research should use those topologies as a baseline~\cite{rodan2011minimum}\cite{maat2018optimization}.
The last examined topology is called permutation and it is created by randomly shuffling the rows of the ring topology matrix.
This operation effectively splits the long ring to a number of smaller rings of different lengths.

Some authors (e.g.,~\cite{maat2018optimization}) use the same constant for all the nonzero connection weights in the ring, chain, and permutation topologies instead of generating the values from $\text{N}(\mu_{res}, \sigma_{res}^2)$ as is the case for the sparse topology (e.g.,~\cite{gallicchio2019topologies}).
In other words, the reservoir matrix $W$ can be expressed as $\lambda W_b$, where $W_b$ is a binary matrix and $\lambda$ is the desired constant.
During our experiments, we have avoided this restriction in order not to give an unfair advantage of an extra degree of freedom to the sparse topology.
All the weights are generated from $\text{N}(\mu_{res}, \sigma_{res}^2)$ and both $\mu_{res}$ and $\sigma_{res}$ are part of the optimized hyperparameters.
If favorable, the optimization algorithm can enforce the constant values by setting $\mu_{res}=\lambda$ and $\sigma_{res}=0$.

It is worth noting that the sparse topology has $\text{O}(n^2)$ parameters where $n$ is the number of neurons, whereas the ring, chain and permutation topologies have only $\text{O}(n)$ parameters. Analogously to other papers (e.g.,~\cite{gallicchio2019topologies}~\cite{rodan2011minimum}), we compare topologies with the same number of neurons, not the same number of parameters.

\subsection{CMA-ES}

Covariance Matrix Adaptation Evolution Strategy (CMA-ES~\cite{hansen2001cma}) is a continuous optimization algorithm from a wide class of evolutionary strategies. The algorithm generates a population of candidate solutions, evaluates their fitness and estimates the natural gradient. Afterwards, it makes a step in the direction of the gradient in order to minimize the fitness. During the evolution, the algorithm adapts the covariance matrix from which it generates the population.

ESNs do not seem to have a single optimal set of hyperparameters, but instead, the parameters must be in a fragile balance. Therefore, CMA-ES is well suited for the task of ESN hyperparameter optimization as the covariance matrix captures the dependencies between individual hyperparameters and generates sensible candidates.

In our experiments, we use a modification called \textit{active CMA-ES}~\cite{jastrebski2006improving}, which updates its covariance matrix based on both successful and unsuccessful candidates, therefore avoiding unpromising parts of the search space.
Thorough explanation of CMA-ES is beyond the scope of this paper, please refer to, e.g., the tutorial paper by \citeauthor{hansen2016tutorial}~\cite{hansen2016tutorial} for more details.

\subsection{Ordered and Chaotic Dynamics}

Let us provide a brief informal description of chaotic and ordered dynamics.
In an \textit{ordered system}, a small perturbation of initial conditions tends to vanish with time.
In a \textit{chaotic system}, on the other hand, even the smallest perturbation amplifies and eventually leads to a very different behavior of the system.

Echo state networks depending on their hyperparameters can be anywhere between total order and absolute chaos (Figure~\ref{fig:three_dynamics}). In between those two regimes, there lies a narrow band called the \textit{edge of chaos}. Some authors claim that the ordered side of the edge of chaos is the setting where the echo state networks demonstrate the best performance and maximize various information theory measures~\cite{matzner2017neuroevolution}\cite{bertschinger2004real}.

For more details regarding the chaos theory and time series analysis in general, please refer to~\citeauthor{sprott2003chaos}~\cite{sprott2003chaos}.

\subsection{Benchmarks}

The selected topologies will be optimized to maximize the performance on two univariate time series prediction benchmarks. Unfortunately, researchers have not yet converged to a single and easily comparable performance measure and even though there exist widely used benchmark tasks, many authors have developed their specific modification or parametrization. Unless specified otherwise, we will use the measures from \citeauthor{gallicchio2019topologies}~\cite{gallicchio2019topologies}.

\subsubsection{NARMA10}

In the first task the network is driven by a random input sequence and its target is a \textit{Non-linear Autoregressive Moving Average} of 10th order (NARMA10), i.e., a nonlinear combination of the last 10 inputs and outputs. Formally, the desired output sequence is defined as follows:
\begin{equation*}
    \textstyle
    y(t+1) = 0.3y(t) + 0.05y(t) \sum_{i=0}^{9} y(t-i) + 1.5u(t-10)u(t) + 0.1 \,,
    \label{eq:narma10}
\end{equation*}
where $u(t)$ is the random input sequence generated from uniform distribution $U(0, 0.5)$ in time $t$ and $y(t)$ is the desired output sequence in time $t$.

\begin{figure}
	\centering
	\includegraphics[width=\linewidth]{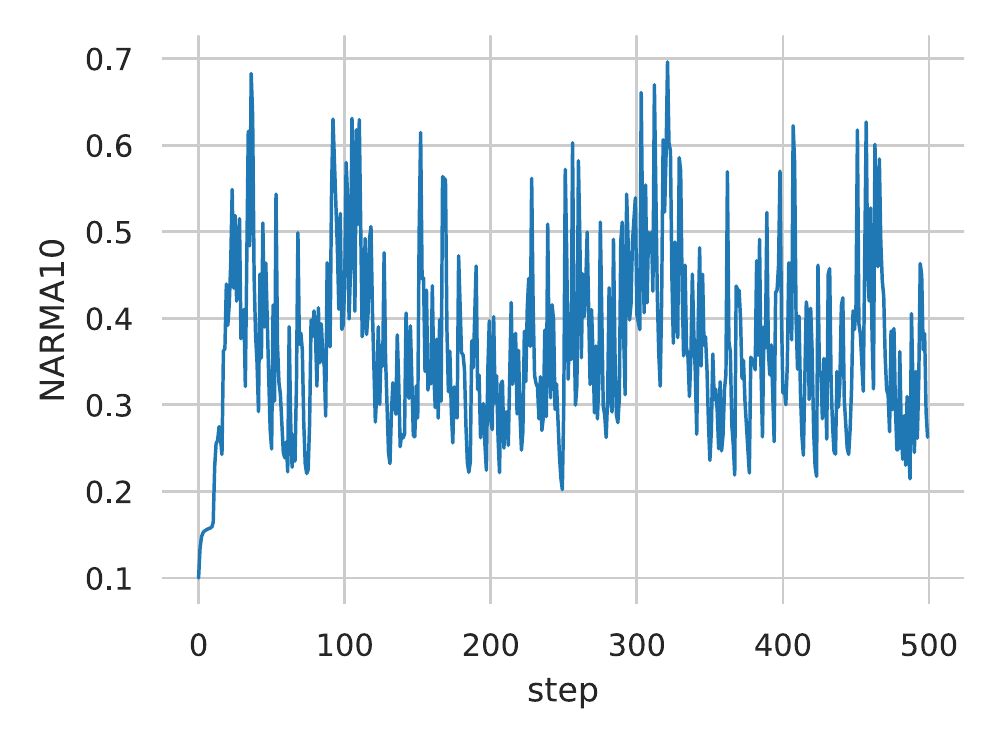}
	\caption{A possible realization of the NARMA10 sequence.}
	\label{fig:narma10}
	\Description{Chaotic plot of the possible NARMA10 sequence.}
\end{figure}

Recent research by \citeauthor{kubota2019narma}~\cite{kubota2019narma} identified that NARMA10 sequence has a significant probability of divergence. Even though it makes it an inconvenient comparison benchmark, it is still widely used in ESN literature, both historical and new. To prevent the divergence in our experiments, we check whether the generated NARMA10 sequences are bounded by $[-1, 1]$ and regenerate them otherwise. One possible realization of the NARMA10 sequence is demonstrated in Figure~\ref{fig:narma10}.

\subsubsection{Mackey-Glass Equation}

\begin{figure}
	\centering
	\includegraphics[width=\linewidth]{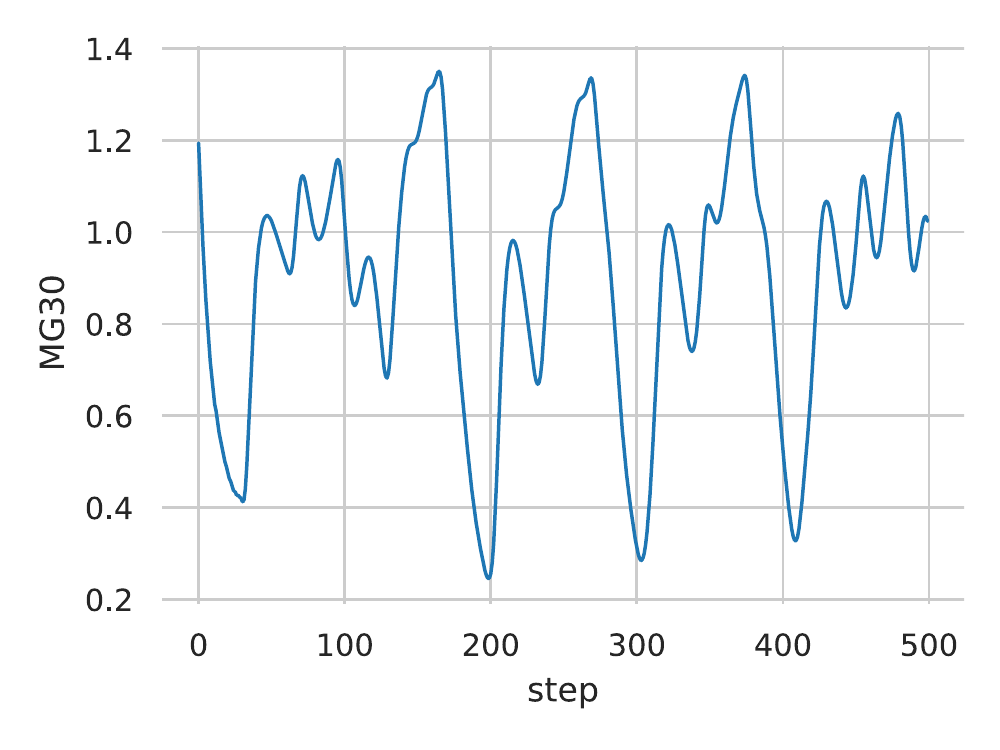}
	\caption{A possible realization of the MG30 sequence.}
	\label{fig:mg30}
	\Description{Repeating plot of MG30 chaotic attractor.}
\end{figure}

In the second benchmark, the network is driven by a sequence generated from \textit{Mackey-Glass} equations (e.g.,~\cite{jaeger2001echo}) and its task is to predict the next value. The desired sequence is obtained by discretizing the following differential equation:
\begin{equation*}
    \textstyle
    \dfrac{\delta u(t)}{\delta t} = \dfrac{0.2 u(t-\tau) }{1 + u(t-\tau)^{10}} - 0.1u(t)
    \label{eq:mgtau}
\end{equation*}
where $\tau$ is a parameter of the equation, and $\delta t$ is the discretization constant set to $0.1$. Adopting the method from \citeauthor{jaeger2001echo}~\cite{jaeger2001echo}, the discretization is approximated as:
\begin{equation*}
	\textstyle
	u(t+1) = u(t) + \delta \left( \dfrac{0.2 u(t-\frac{\tau}{\delta}) }{1 + u(t-\frac{\tau}{\delta})^{10}} - 0.1u(t) \right)
	\label{eq:mgtauapx}
\end{equation*}
and the resulting sequence is subsampled by taking every tenth element. Furthermore, before feeding the values to the network, they are transformed to interval $[-1, 1]$ by $x \rightarrow \tanh({x - 1}$) and the output of the network is transformed back to the original range by $x \rightarrow \atanh(x) - 1$. Throughout this paper, we use $\tau=30$ and $\tau=17$ and call the benchmarks MG30 and MG17 respectively. One possible realization of the MG30 sequence is demonstrated in Figure~\ref{fig:mg30}.

Unless specified otherwise, the performance of all tasks is measured using the \textit{mean squared error} (MSE) of the desired and the predicted output sequences. In some experiments, we will be using \textit{normalized mean squared error} (NMSE) defined as:
\begin{equation*}
	\textstyle
	NMSE = \sum_{t=1}^{N} \dfrac{ \left( y(t) - x(t) \right) ^2 } { N \sigma^2 } \,,
	\label{eq:nmse}
\end{equation*}
where $y(t)$ is the desired output sequence and $x(t)$ is the predicted sequence, $N$ is the length of the sequences, and $\sigma^2$ is the variance of the desired output sequence. And the last measure is \textit{normalized root mean squared error} defined as:
\begin{equation*}
	\textstyle
	NRMSE = \sqrt{NMSE} \,.
	\label{eq:nrmse}
\end{equation*}

\section{Experimental Environment} \label{sec:experiment}

The hyperparameters of each reservoir topology are optimized so that the instantiated network maximizes its performance on one of the benchmark tasks.
Similarly to~\cite{gallicchio2019topologies}, the experiment uses ESNs with $500$ reservoir neurons, regardless of the topology.
The reservoir weights are generated from normal distribution $\text{N}(\mu_{res}, \sigma_{res}^2)$, feedback weights from uniform distribution $U(-\omega_{fb}, \omega_{fb})$, and input weights from $U(-\omega_{in}, \omega_{in})$.
Every weight in the reservoir will be set to zero with the probability of $\beta$. The optimized hyperparameters are reservoir parameters $\sigma_{res}$ and $\mu_{res}$, input weight spread $\omega_{in}$, feedback weight spread $\omega_{fb}$, sparsity $\alpha$, leakage $\gamma$, bias $\mu_{b}$ and internal noise $\epsilon$.
Parameters $\sigma_{res}$ and $\epsilon$ are optimized in logarithmic space via $x \longmapsto e^{-50x}$.
Parameters $\mu_{res}$, $\omega_{in}$, $\omega_{fb}$, and $\mu_{b}$ are optimized in square root space via $x \longmapsto 2 x |x|$ to make more cautions steps in the close proximity to zero.
The hyperparameters for every topology are optimized ten times over and in each of those trials, the best encountered network is evaluated a hundred times using a freshly generated inputs and outputs. The trial with the best mean performance is reported.

The starting parameters do not seem to be that important as long as the corresponding network is not overly chaotic or overly stable and the optimizer can detect an improving gradient.
We handpicked those to $\sigma_{res}=\frac{1}{2k}$, where $k$ is the average number of neuron inputs, $\omega_{in}=0.02$, $\omega_{fb}=0$, $\beta=0.1$, $\gamma=0.9$, $\mu_{b}=0$, and $\epsilon=4.5e^{-5}$.
The initial sigma for CMA-ES is set to $0.05$ except for $\sigma_{res}$ and $\omega_{fb}$ where we use $0.01$.
The optimization is bounded to interval $[-1.1; 1.1]$ (before the search space transformation) and limited to 5000 evaluations which is more than sufficient to converge for all the tested combinations of topologies and benchmark tasks.

The initial neuron activation is set to zero and the first 1000 steps are used as a washout period to eliminate the effect of state initialization.
The subsequent 5000 steps are used for linear regression training and the last 5000 steps are used for performance test.

In a few experiments where performance comparison may be unclear, we perform significance testing using Welch’s t-test with statistical significance level of $0.05$.

\section{Results}

\begin{table*}
    \centering
    \setlength{\tabcolsep}{15pt}
    \begin{tabular}{lrrr}
        \toprule
        topology & NARMA10 & MG17 & MG30 \\
        \midrule
        sparse        & $\num{4.67e-09} \; \pm \num{3.0e-09}$
                      & $\num{1.27e-13} \; \pm \num{1.8e-14}$
                      & $\num{6.74e-10} \; \pm \num{1.5e-10}$ \\
        permutation   & $\num{6.34e-09} \; \pm \num{1.6e-09}$
                      & $\num{2.40e-13} \; \pm \num{6.0e-14}$
                      & $\num{5.06e-10} \; \pm \num{8.0e-11}$ \\
        chain         & $\num{8.18e-09} \; \pm \num{3.2e-09}$
                      & $\num{1.70e-13} \; \pm \num{2.8e-14}$
                      & $\num{5.54e-10} \; \pm \num{8.4e-11}$ \\
        ring          & $\num{1.28e-08} \; \pm \num{5.9e-09}$
                      & $\num{2.28e-13} \; \pm \num{8.6e-14}$
                      & $\num{6.89e-10} \; \pm \num{1.1e-10}$ \\
        \bottomrule
    \end{tabular}
	\caption{The MSE results of the four tested topologies on NARMA10, MG17 and MG30 benchmarks}
	\label{tab:results_gallancchio}
\end{table*}

\begin{figure}
	\centering
	\includegraphics[width=\linewidth]{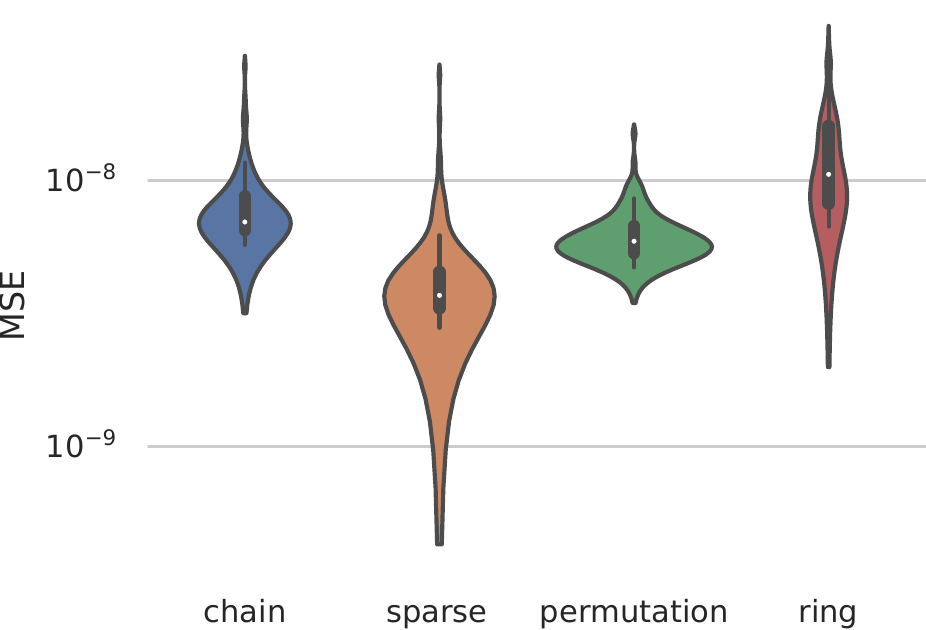}
	\caption{MSE on the NARMA10 task.}
	\label{fig:narma10_gallancchio}
	\Description{Violin plot representing the distribution of results on the NARMA10 task for each topology.}
\end{figure}

The results for the NARMA10 task are shown in Figure~\ref{fig:narma10_gallancchio} and Table~\ref{tab:results_gallancchio}. The sparse topology slightly outperformed all the other topologies ($p<0.05$).
The presented results differ by multiple orders of magnitude from e.g. \citeauthor{gallicchio2019topologies}~\cite{gallicchio2019topologies}, who reached NARMA10 $\approx 10^{-4} \pm 10^{-5}$, MG17 $\approx 10^{-9} \pm 10^{-10}$, and MG30 $\approx 10^{-8} \pm 10^{-9}$.
The parameters in the aforementioned work were tuned manually and the sparse topology ended up as the worst of the four.

\begin{table}
	\centering
	\setlength{\tabcolsep}{15pt}
	\begin{tabular}{lr}
		\toprule
		topology & NARMA10 (no feedback)\\
		\midrule
		sparse        & $\num{7.0e-5} \; \pm \num{3.9e-5}$
		\\
		permutation   & $\num{5.1e-5} \; \pm \num{2.6e-5}$
		\\
		chain         & $\num{5.9e-5} \; \pm \num{2.6e-5}$
		\\
		ring          & $\num{7.1e-5} \; \pm \num{7.2e-5}$
		\\
		\bottomrule
	\end{tabular}
	\caption{The MSE results of the four tested topologies without feedback connections on NARMA10 benchmarks.}
	\label{tab:results_nofb_narma10_gallancchio}
\end{table}

\begin{figure}
	\centering
	\includegraphics[width=\linewidth]{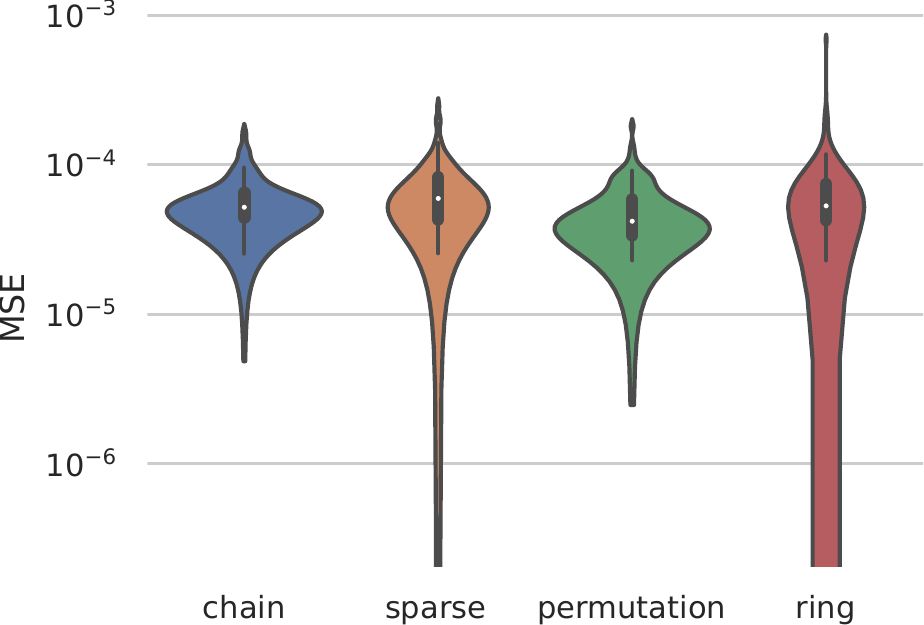}
	\caption{MSE on the NARMA10 task for networks without feedback connections.}
	\label{fig:nofb_narma10_gallancchio}
	\Description{Violin plot representing the distribution of results on the NARMA10 task for each topology without feedback weights.}
\end{figure}

Most of the difference from~\cite{gallicchio2019topologies} is caused by the fact that the authors have not used feedback connections, which even though difficult to tune manually~\cite{koryakin2012balanced} can staggeringly improve the results in autoregressive tasks, e.g., NARMA10.
Furthermore, ESNs with feedback achieve universal computational capabilities~\cite{maass2005universal}.
Note that there is no benefit in using feedback connections for tasks where the input and the desired output sequences are equal and only shifted by one step (e.g., our variants of MG17 and MG30).
In those tasks, the network is already fed by the desired output and feeding its own prediction through feedback would only introduce a new kind of noise for which the network has not been trained \footnote{During the training, the network is fed by the desired output using the teacher-forced signal.}.

To demonstrate the effect of feedback connections, the NARMA10 experiment was repeated without the feedback connections as well and the results can be found in Figure~\ref{fig:nofb_narma10_gallancchio} and Table~\ref{tab:results_nofb_narma10_gallancchio}. The results are orders of magnitude worse compared to the results with feedback connections, however, there is still a significant improvement compared to the results presented in~\cite{gallicchio2019topologies} and all of the topologies still perform similarly.
The authors in~\cite{gallicchio2019topologies} propose a hypothesis that permutation, ring and chain topologies provide better performance than the sparse topology on NARMA10, MG17 and MG30 benchmarks. Our results oppose the hypothesis by properly tuning the hyperparameters of the sparse topology.
Furthermore, the CMA-ES results on the aforementioned tasks surpassed even multi-layered \textit{Deep ESN} results from the discussed paper.

With a slight modification of the experiments, the results can be compared also with \textit{swarm optimization} method by \citeauthor{wang2015swarm}~\cite{wang2015swarm}.
In NARMA10 task with a network counting 500 neurons, the CMA-ES procedure reached NMSE $\num{5.2e-7} \pm \num{1.95e-4}$ compared to $0.0346$ reported in~\cite{wang2015swarm}.
In MG84-17 task, the $\text{NRMSE}_{84}$ is $\num{1.28e-4} \pm \num{4.8e-5}$ for the CMA-ES and $\num{1.4e-3}$ for swarm optimization. It is worth mentioning that the authors have reached the results in a fraction of function evaluations compared to our method, however, the progress of the swarm optimization stalled after $\approx 150$ iterations.

Further comparison can be done e.g., with the benchmark variants by \citeauthor{cerina2020echobay} \cite{cerina2020echobay}.
Adapting the measures defined in the paper, our method reached NRMSE in NARMA10~$\approx 99.928\%$, and MG10-17~$\approx 99.993\%$.
The presented results were NARMA10~$\approx 90.647\%$ and MG10-17~$\approx 99.98\%$.
It should be pointed out that the goal of~\cite{cerina2020echobay} was to create a constrained network for devices with limited computational and memory resources, not to solely maximize the performance measures.

In comparison with another paper discussing different topologies by \citeauthor{cernansky2008predict}\cite{cernansky2008predict}, our CMA-ES procedure reached MG17 MSE~$=\num{6.49e-16} \pm \num{3.99e-17}$ and NARMA10 MSE~$=\num{1.4e-4} \pm \num{7.4e-5}$ versus the best presented results in the paper MG17 MSE~$=\num{9.82e-16} \pm \num{9.02e-17}$ and NARMA10 MSE~$=\num{8.4e-4} \pm \num{8.4e-5}$. Even though the results presented in \cite{cernansky2008predict} are slightly surpassed by our method, the authors have reached impressive performance simply by dividing the reservoir matrix by its largest eigenvalue. This scaling technique transforms the reservoir to be close to the edge of chaos dynamics. \footnote{It should be noted that the authors use a large network of 1000 neurons for the MG17 task and even better results can be achieved by increasing the training sequence length.}

\begin{table}
	\centering
	\setlength{\tabcolsep}{15pt}
	\begin{tabular}{lr}
		\toprule
		topology & NARMA10 \\
		\midrule
		sparse        & $\num{3.08e-4} \; \pm \num{1.7e-4}$
		\\
		permutation   & $\num{1.19e-3} \; \pm \num{7.6e-4}$
		\\
		chain         & $\num{1.49e-3} \; \pm \num{9.6e-4}$
		\\
		ring          & $\num{1.79e-3} \; \pm \num{2.6e-3}$
		\\
		\bottomrule
	\end{tabular}
	\caption{The NMSE results of the four tested topologies on NARMA10 variant by \citeauthor{rodan2011minimum}.}
	\label{tab:narma10_rodan}
\end{table}

\begin{figure}
	\centering
	\includegraphics[width=\linewidth]{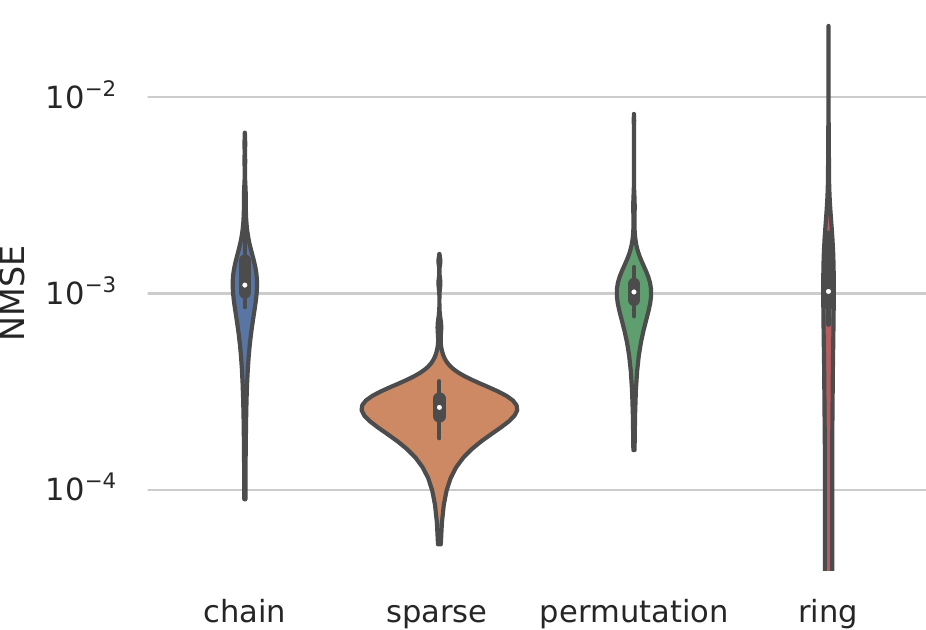}
	\caption{NMSE on the NARMA10 variant by \citeauthor{rodan2011minimum}~\cite{rodan2011minimum}.}
	\label{fig:narma10_rodan}
	\Description{Violin plot representing the distribution of results on the NARMA10 variant by \citeauthor{rodan2011minimum}~\cite{rodan2011minimum}.}
\end{figure}

Our method is also superior to the topology comparison results by \citeauthor{rodan2011minimum}\cite{rodan2011minimum}. The authors of this paper were one of the first to introduce the idea of minimum complexity echo state networks, specifically variants of the ring topology. They conclude that the proposed simple topologies offer a similar performance to the sparse topology while being easier to construct and easier to train. We managed to reproduce and significantly improve their results in NARMA10 task (reported NMSE $\approx 0.04$) with reservoir size 200 for the chain, ring, and permutation topologies. However, the sparse topology significantly outperformed all of them ($p<0.05$). See Figure~\ref{fig:narma10_rodan} and Table~\ref{tab:narma10_rodan}.

\citeauthor{holzmann2009filter}\cite{holzmann2009filter} experimented not with a topology, but with the neuron itself. They designed a neuron with band pass filter and a new kind of delay\&sum readout. The results were compared with a plain ESN on a more difficult version of the NARMA10 benchmark with lag variable $\tau$. In their paper, the plain ESN error quickly soars as the $\tau$ increases and for $\tau>2$ its NRMSE reaches values up to $0.8$, while their improved version of the network stays just slightly above $0.4$. We have reproduced their results with our CMA-ES optimization and the plain ESN results outperformed all of the modifications. For $\tau=2$, NRMSE in the NARMA10 task was $0.36 \pm 0.01$ and for $\tau=4$, it was $0.38 \pm 0.01$.

\begin{table}
	\centering
	\setlength{\tabcolsep}{15pt}
	\begin{tabular}{lrr}
		\toprule
		reservoir size & CMA-ES & GA \\
		\midrule
		200   & $\num{0.12} \; \pm \num{0.005}$ & $\num{0.15} \; \pm \num{0.005}$
		\\
		400   & $\num{0.02} \; \pm \num{0.001}$ & $\num{0.09} \; \pm \num{0.013}$
		\\
		\bottomrule
	\end{tabular}
	\caption{The NMSE results on the NARMA30 task by \citeauthor{dale2018neuroevolution}~\cite{dale2018neuroevolution}.}
	\label{tab:narma30_dale}
\end{table}

Finally, the CMA-ES results with plain ESN are superior to a hierarchy of ESNs optimized by a genetic algorithm on NARMA30 task by~\citeauthor{dale2018neuroevolution}~\cite{dale2018neuroevolution} with results presented in Table~\ref{tab:narma30_dale}.

\section{Discussion} \label{sec:discussion}

According to the results, plain ESN with the sparse reservoir topology can achieve extraordinary performance as long as it has been properly tuned. Unfortunately, many authors appear to underestimate hyperparameter optimization and compare their innovations with subpar ESN instances. To their defense, manual parameter tuning is very difficult and grid search cannot cover the small nuances created by fragile recurrent dynamics.

\begin{figure}
	\centering
	\includegraphics[width=\linewidth]{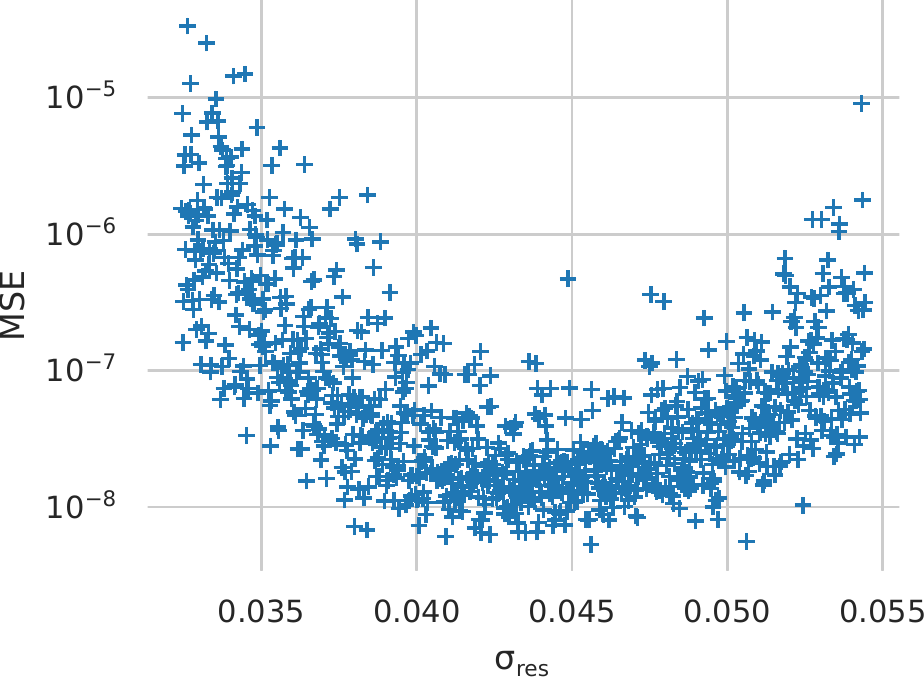}
	\caption{MSE versus $\sigma_{res}$ in a narrow range around the optimum.}
	\label{fig:sigma_res}
	\Description{Scatter plot demonstrating the sensitivity to slight parameter changes for $\sigma_{res}$.}
\end{figure}

Let us demonstrate an example. Taking the hyperparameters of one of the best performing sparse networks on the NARMA10 benchmark, let us plot a few perturbations close to the optimal $\sigma_{res}$ in Figure~\ref{fig:sigma_res}. The optimal interval is very narrow, for instance, when $1e-2$ is subtracted from the $\sigma_{res}$, the error on the NARMA10 task soars by more than one order of magnitude.

\begin{figure}
	\centering
	\includegraphics[width=\linewidth]{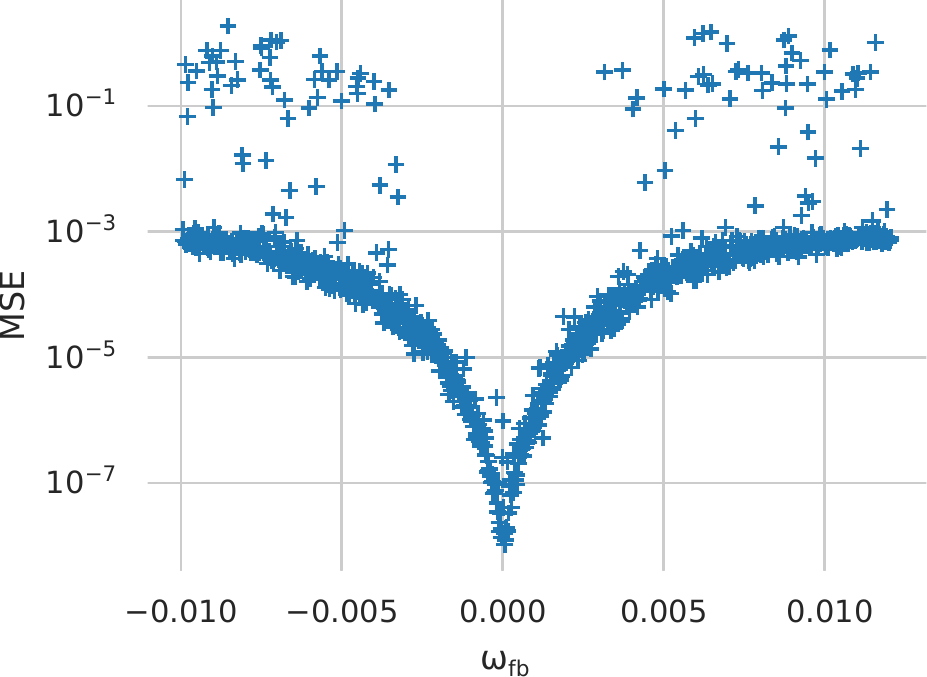}
	\caption{MSE versus $\omega_{fb}$ on a coarse grid.}
	\label{fig:fb_weight}
	\Description{Scatter plot demonstrating the sensitivity to slight parameter changes for $\omega_{fb}$.}
\end{figure}

\begin{figure}
	\centering
	\includegraphics[width=\linewidth]{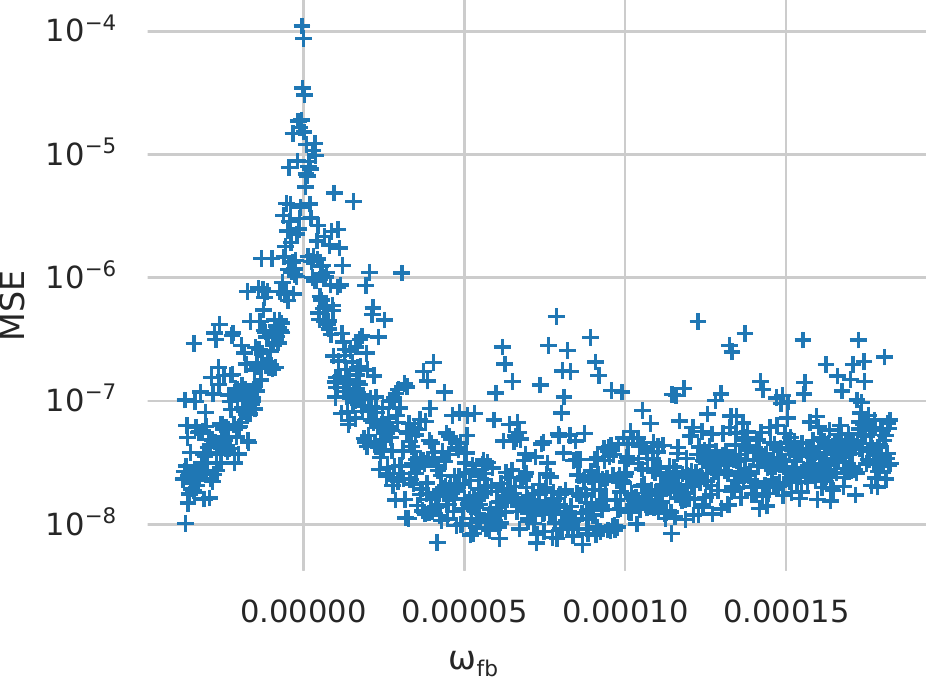}
	\caption{MSE versus $\omega_{fb}$ in a narrow range around the optimum.}
	\label{fig:fb_weight_detail}
	\Description{Scatter plot demonstrating the sensitivity to slight parameter changes for $\omega_{fb}$ on a denser grid.}
\end{figure}

To further demonstrate the pitfalls of hyperparameter tuning, Figure~\ref{fig:fb_weight} shows a scatter plot with the effect of small perturbations of $\omega_{fb}$. The required grid density would need to be $\approx \num{1e-3}$ to recognize the depicted pattern. By the shape of the plot, one could assume that feedback weights are inadequate for the task as the error is minimized when $\omega_{fb}$ is set to zero. Unfortunately, if we plot a detail with even denser grid of step size at most $\approx \num{1e-5}$ as shown in Figure~\ref{fig:fb_weight_detail}, we can clearly see that feedback weights are necessary. However they need to be set to a very narrow range between $\num{5e-5}$ and $\num{1e-4}$.

During our experiments, we have encountered situations where this steep decrease in performance in the vicinity of zero blocked the CMA-ES algorithm from changing the sign of the optimized variable. This poses a problem especially in later stages of the evolution, when the spread of the candidates becomes smaller. We recommend running the optimization multiple times to increase the probability of obtaining the correct sign for all the variables.

Considering the necessity of a very dense hyperparameter grid and the fact that there are eight parameters to be optimized, grid search becomes nearly impossible due to the \textit{curse of dimensionality}. What is convenient about the restricted topologies, such as ring or chain, is that each neuron has a smaller number of input connections and the network has a stronger tendency to stay in the ordered regime. Therefore, the range of hyperparameters in which the network provides reasonable performance dilates, and thus becomes easier to optimize manually or via grid search.

\begin{figure}
	\centering
	\includegraphics[width=\linewidth]{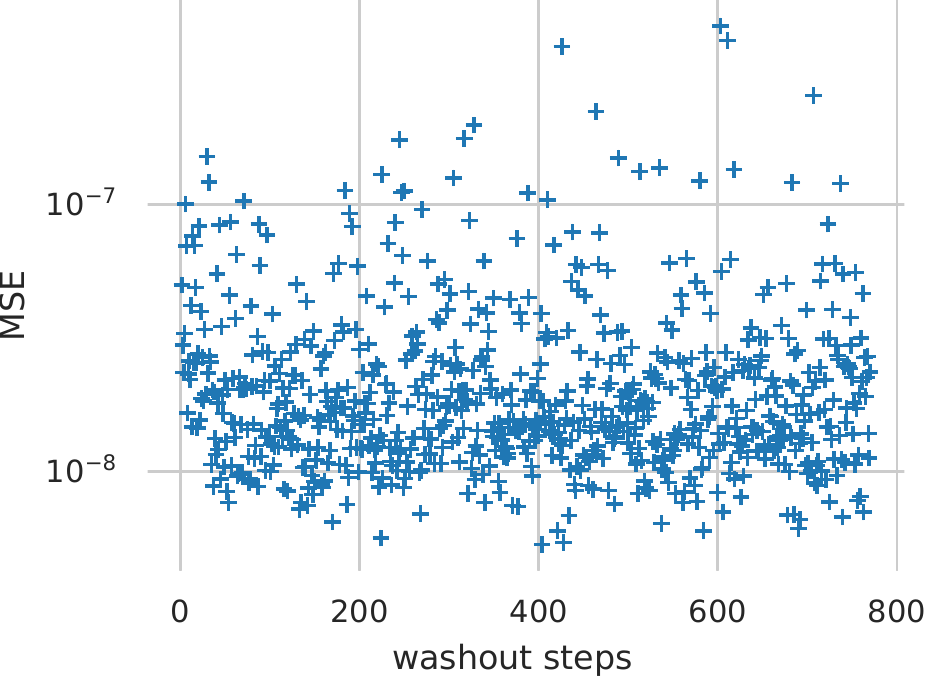}
	\caption{MSE versus the number of washout steps.}
	\label{fig:washout_steps}
	\Description{Scatter plot demonstrating the influence of the number of washout steps to the final performance.}
\end{figure}

\begin{figure}
	\centering
	\includegraphics[width=\linewidth]{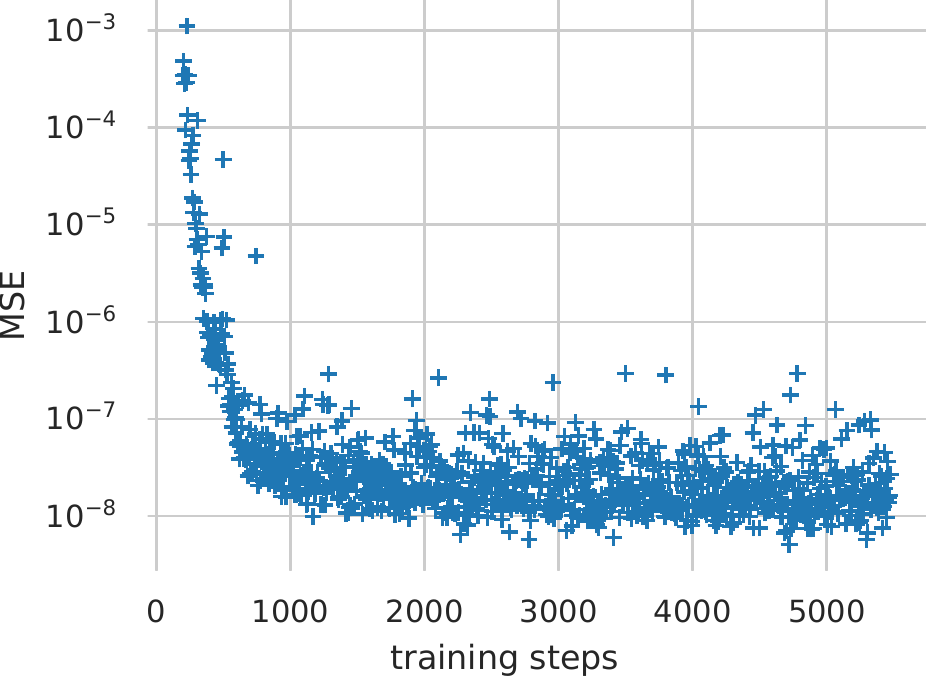}
	\caption{MSE versus the number of training steps.}
	\label{fig:training_steps}
	\Description{Scatter plot demonstrating the influence of the number of training steps to the final performance.}
\end{figure}

One thing we have seen to vary in the literature are the lengths of the washout and training sequences. A rule of thumb is to have at least as many washout steps as there are neurons in the network. The necessary number of washout steps may depend on the topology of the network (specifically its diameter) and its dynamics, especially the reach of its memory. For instance, in our NARMA10 network, anything above 100 seems sufficient (Figure~\ref{fig:washout_steps}). What is more demanding is the train sequence length. As depicted in Figure~\ref{fig:training_steps} our network requires at least 4000 training steps to reach its full potential.

\begin{figure}
	\centering
	\includegraphics[width=\linewidth]{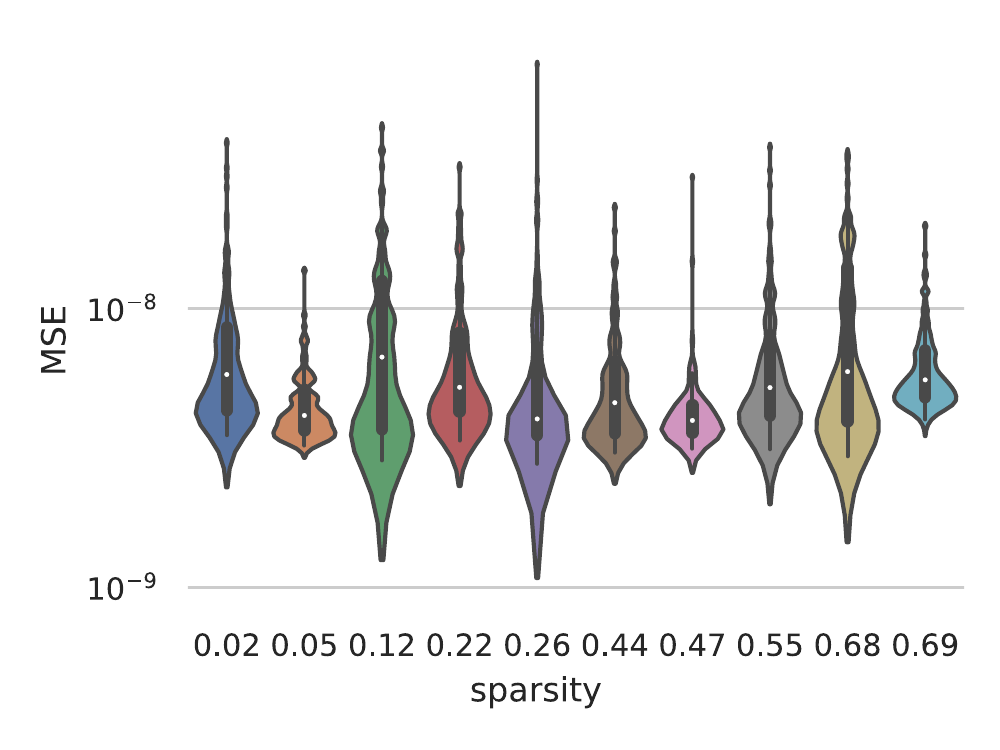}
	\caption{The performance of the best individuals from ten independent evolutionary runs on the NARMA10 task plotted against their sparsity.}
	\label{fig:sparsity_correlation}
	\Description{Violin plot showing that the performance of the best individual is not correlating with sparsity.}
\end{figure}

Another noteworthy observation is that the sparsity parameter $\beta$ in the sparse topology did not seem to play an important role and the best found sparse networks balanced other parameters accordingly without a noticeable performance loss (Figure~\ref{fig:sparsity_correlation}).
However, it should be pointed out that none of the ten networks has sparsity above 70\%.
Nonetheless, the likely reason is that there is simply no gradient pushing the sparsity higher from the initial value of 10\% and the sparsity parameter merely performs a random walk around its origin.
Naturally, the sparsity plays much more important role in the circle and chain topologies, where each dropped connection creates a new disconnected component.
The sparsity of the best individuals with those two topologies has always been optimized out to zero.

During our experiments, we have encountered many extreme values, especially while evaluating the ring and chain topologies.
One explanation could be that those topologies may be prone to generating a subpar reservoir instance, e.g., a ring where a few connections gets a very low weight and, consequently, create a bottleneck not passing the data further in the ring.
Taking this property into account when using ESNs in practice, one should always train multiple instances and use the best performing one.

\section{Conclusion}

We have investigated the challenge of hyperparameter tuning in echo state networks with multiple different topologies.

Our results suggest that restricted topologies, such as chain, ring and permutation do not provide any significant performance edge over a plain sparse network with properly tuned hyperparameters and even underperform in some of the evaluated tasks.

We argue that grid search is not a proper way to tune hyperparameters because a small perturbation can cause large differences in performance and the grid would need to be impractically dense.
This is especially true for networks close to the edge of chaos.

Results of papers comparing various ESN innovations, such as topologies, are questionable without a proper hyperparameter tuning.
For trustworthy evaluation of novelties in ESNs, we propose the usage of CMA-ES as a robust, yet computationally intensive tool to tune the hyperparameters.

To the best of our knowledge, the proposed results form a new baseline for echo state networks in NARMA10, MG17, and MG30 benchmarks for ESNs with the corresponding number of neurons.

We have implemented a highly optimized C++ framework for ESN optimization with GPU support. The source code will be publicly available on our GitHub page \cite{github2022echo} under permissive license including the configuration for all the evaluated experiments.

\section{Citations and Bibliographies}

\begin{acks}
This research was supported by Charles University grant SVV-260588 and GA UK project number 1578717. Computational resources were supplied by the project ``e-Infrastruktura CZ'' (e-INFRA CZ LM2018140) supported by the Ministry of Education, Youth and Sports of the Czech Republic. We thank the authors of ArrayFire~\cite{arrayfire} and libcmaes~\cite{libcmaes} software packages for sharing their hard work under open source licences.
\end{acks}

\bibliographystyle{ACM-Reference-Format}
\bibliography{bibliography}

\appendix

\end{document}